\documentclass[10pt]{article}

\usepackage{microtype}
\usepackage{graphicx}
\usepackage{subfigure}
\usepackage{booktabs} 
\usepackage{times}
\usepackage{epsfig}
\usepackage{graphicx}
\usepackage{amsmath}
\usepackage{amssymb}
\usepackage{multirow}
\usepackage{algorithm,algorithmic}
\usepackage[accepted]{icml2021}

\usepackage{hyperref}
\hypersetup{
    colorlinks=true,
    linkcolor=blue,
    citecolor=blue,
    urlcolor=blue,
}

\icmltitlerunning{stochastic MG/OPT}
\begin{document}
\twocolumn[
\icmltitle{Training of deep residual networks with stochastic MG/OPT}
\icmlsetsymbol{equal}{*}

\begin{icmlauthorlist}
\icmlauthor{Cyrill von Planta}{to}
\icmlauthor{Alena Kopani\v{c}\'akov\'a}{to}
\icmlauthor{Rolf Krause}{to}
\end{icmlauthorlist}

\icmlaffiliation{to}{Euler Institute, Università della Svizzera Italiana}
\icmlcorrespondingauthor{Cyrill von Planta}{cyrill.von.planta@usi.ch}
\icmlkeywords{Machine Learning, Multilevel optimization, FAS, MG/OPT}
\vskip 0.3in
]

\printAffiliationsAndNotice{} 


\date{Arxiv version of ICML2021 workshop minipaper from August 9, 2021}

\begin{abstract}
We train deep residual networks with a stochastic variant of the nonlinear multigrid method MG/OPT. To build the multilevel hierarchy, we use the dynamical systems viewpoint specific to residual networks. We report significant speed-ups and additional robustness for training MNIST on deep residual networks. Our numerical experiments also indicate that multilevel training can be used as a pruning technique, as many of the auxiliary networks have accuracies comparable to the original network.
\end{abstract}

\section{Introduction}
\label{sec_intro}

Deep residual networks (ResNets)~\cite{he2016deep} are the state-of-the-art architecture for a variety of computer vision tasks \cite{lin2014microsoft,le1989handwritten,russakovsky2015imagenet}.
The key innovation behind ResNets is the residual block, which allows information to be passed directly through, making the backpropagation less prone to exploding or vanishing gradients. 
This enabled the training of networks with hundreds of layers, which in turn led to significant gains in prediction power.

The processing and memory costs to train these networks with backpropagation dramatically increase with the size of the network.
To apply networks in  low-end devices, pruning~\cite{tung2018deep, zhu2017prune} and sparsity techniques~\cite{changpinyo2017power, han2016dsd} are used to remove a significant fraction of the network weights, while preserving test accuracy attained by full models. 
Pruning methods can be computationally expensive, thus algorithms that produce light-weight models without extra cost during training are well sought.

To address these issues, in this article we employ a variant of the multigrid method.
The multigrid method \cite{Fed62, Hac13} is well known due to its ability to solve linear elliptic problems with optimal complexity.
In machine learning (ML) we have to minimize non-convex and nonlinear loss functions, hence we adapt a nonlinear multigrid variant to the stochastic setting.
This method is MG/OPT \cite{Nas00}, which itself is a variant of the full approximation scheme (FAS) \cite{Bra77}. 
By using the deterministic method MG/OPT to minimize the loss function over mini-batches, we obtain the method s(tochastic)MG/OPT, i.e. (sMG/OPT).

To use multigrid methods in ML, we need to set up a suitable multilevel hierarchy for the neural network to be trained.
To this end, we use the dynamical systems viewpoint~\cite{CMH+17,HRH+18,HR17,Wei17}, which interprets the forward propagation through a ResNet as a forward Euler scheme of an ordinary differential equation (ODE).

\section{Formulation}
\label{sec_formulation}
We denote with ${\mathcal{D}=\{x^i,c^i\}_{i=1,\ldots,S}}$ the data set with $S$ inputs $x^i \in \mathbb{R}^{n^I}$, $n^I$ being the dimension of the input, and labels $c^i \in \mathbb{R}^{n^c}$, where $n^c$ is the number of classes. We set $x=(x^1, \ldots, x^S)$ and ${c=(c^1, \ldots, c^S)}$, respectively.
The response $y^i \in \mathbb{R}^{n^c}$ of a neural network is denoted with the function $\mathcal{F}$, such that $y^i=\mathcal{F}(\theta,x^i)$, whereby $\theta \in \mathbb{R}^K$ stands  for the entirety of the (flattened) network parameters.

For the loss $\ell(\theta; x,c)$ we use the cross-entropy loss function with $L^2$-regularization.
The learning problem is then defined as finding a set of parameters $\theta^* \in \mathbb{R}^K$, which minimizes the loss, i.e., $ \theta^* = \underset{\theta}{\mathrm{argmin} }\,\ell(\theta; x,c).$ 

Throughout this article we consider ResNets with $N$ layers or ResNet blocks, whereby the state of  each layer is denoted by $y_n$, for  $n=1,\ldots,N$.
The ResNets have fixed widths~$w$, i.e., ${y_n  \in \mathbb{R}^w, \forall n}$.
The first and last layer receive appropriate mappings ${g_F:\mathbb{R}^{n^I} \mapsto \mathbb{R}^w}$ and ${g_L:\mathbb{R}^{n^w} \mapsto \mathbb{R}^{n^c}}$, such that ${y_0^i=g_F(x^i)}$, and  ${y^i = g_L( y^{N})}$, respectively.

At each layer, the state~$y_n$ is given as the response of the $n$-th residual block $\mathcal{F}_n$ and the shortcut connection by
\begin{equation}
\label{eq_res_blk}
y_n = y_{n-1} + \mathcal{F}_n(\theta_n,y_{n-1}),
\end{equation}
where $\theta_n = (W_n,b_n)$, $W_n \in \mathbb{R}^{w \times w}$, $b_n \in \mathbb{R}^{w}$. $\mathcal{F}_n$ is defined as:
\begin{equation}
\mathcal{F}_n(\theta_n,y_{n-1}) =  \textrm{ReLU}(W_n y_{n-1} + b_n).
\end{equation}

\subsubsection*{Dynamical systems viewpoint}

The dynamical systems viewpoint uses the equivalence between the forward propagation of ResNets and the explicit Euler scheme for ODEs.
Given the time interval $[0,T]$ discretized into $N+1$ pieces, the explicit Euler scheme for an ODE $\dot{y}(t) = f(t,y(t)) + \textrm{const},$ has the form 
\begin{equation}
\label{eq_forward_euler}
y_{n+1} = y_n + \Delta_t f(t_n, y_n),\quad  n=1,\ldots, N := \frac{T}{\Delta t},
\end{equation}
where $\Delta_t$ is the time discretization parameter, ${t_n := t_0 + n \Delta_t}$, and ${y_n:= y(t_n)}$. 

\begin{figure}
    \centering
    \includegraphics[width=0.96\columnwidth]{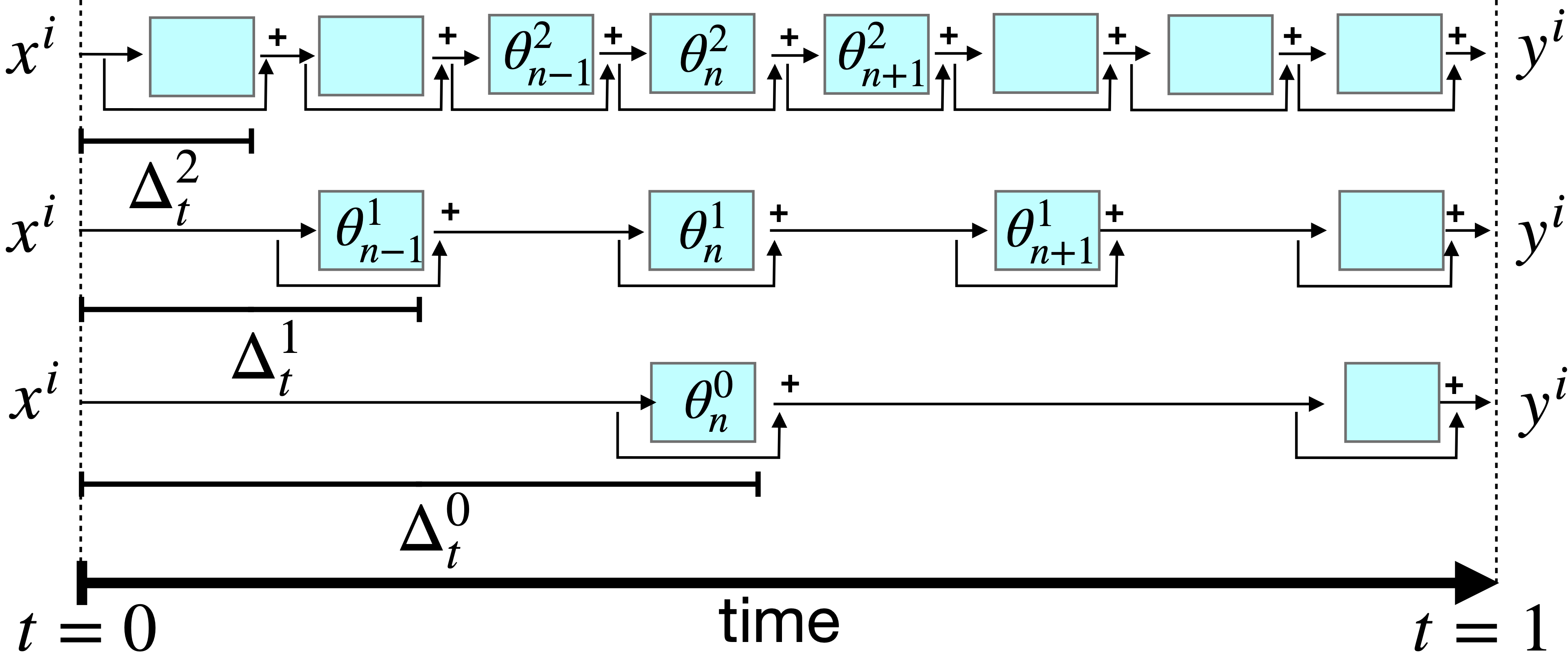}
    \caption{Architecture of three ResNets with $L=2$ and ${\Delta_t^l = 2 \Delta_t^{l+1}}$.}
    \label{fig_ml_ResNet}
\end{figure}

Using the dynamical systems viewpoint, the learning problem for ResNets can be formulated as: Find a ${\theta^* = (\theta_0, \ldots, \theta_N)}$, such that
\begin{align}
\theta^* &= \underset{{\theta}}{\mathrm{argmin} }\,\ell({\theta}; x,c),  \nonumber  \\
\textrm{subject to} \ \  y^i_0 &= g_F(x^i), \quad \forall i=1, \ldots, S, \label{eq_learning_problem}  \\ 
 y^{i}_n &= y^{i}_{n-1} + \Delta_t \; \mathcal{F}_n(y^{i}_{n-1}, \theta_{n}), \; \forall n=1,\ldots,N.  \nonumber
\end{align}
Thus, we can select different values for the discretization parameter~$\Delta_t$ to obtain ResNets of different depths to form a (multilevel) hierarchy (Fig.~\ref{fig_ml_ResNet}), and $\Delta_t$ becomes a scaling parameter for each level.

\subsubsection*{Stochastic MG/OPT}
sMG/OPT is an optimizer to train neural networks by applying the deterministic multilevel method MG/OPT over each mini-batch, the same way SGD applies a gradient descent (GD) step over each mini-batch.
Multilevel methods are iterative solvers, whereby the problem is discretized at different resolutions, giving rise to a multilevel hierarchy. 
The different levels are connected with transfer operators called restriction, (fine to coarse), and prolongation (coarse to fine).   
On each level the solution is smoothed by a smoother, which is normally a simple iterative solver like Jacobi or Gauss-Seidel.
Thus, to set up a multilevel method, we need to specify the multilevel hierarchy, the transfer operators and the smoother.

\begin{figure}
    \centering
    \includegraphics[width=0.96\columnwidth]{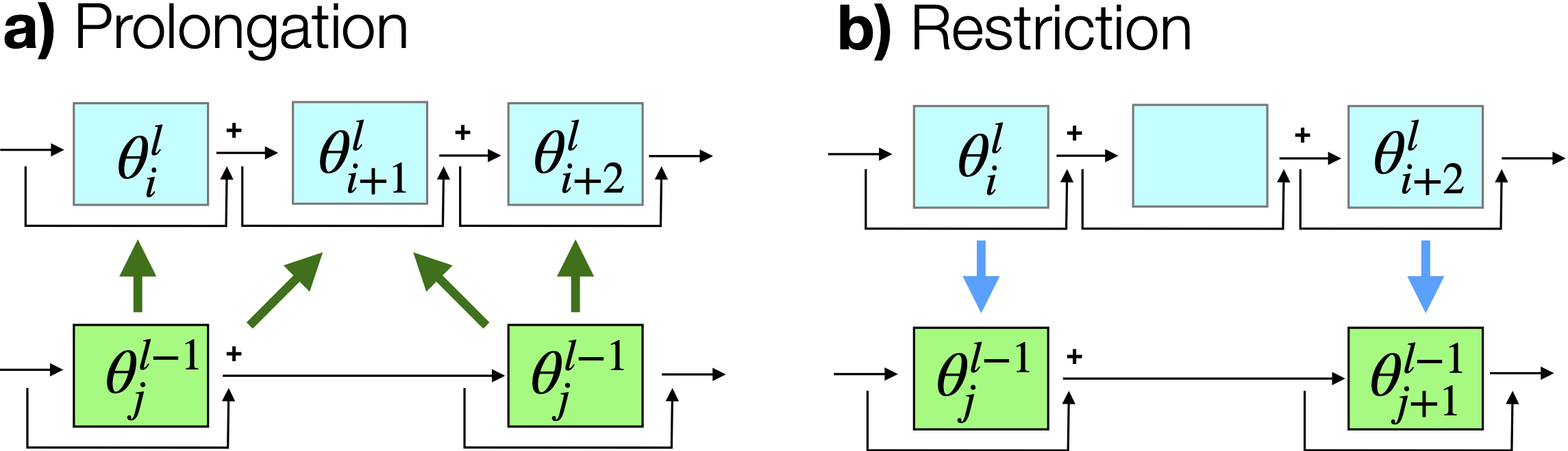}
    \caption{Transfer operators. a) Prolongation. b) Restriction.}
    \label{fig_transferops}
\end{figure}

We create the multilevel hierarchy by assigning each level $l$ a different time step $\Delta^l_t$.
By doubling its size for each level, we generate a hierarchy of ResNets as shown in Fig.~\ref{fig_ml_ResNet}. We set $l=0$ for the coarsest level, and $l=L$ for the finest level.
To transfer information from  finer to coarser levels and vice versa, we take inspiration from algebraic multigrid \cite{Stu01} and interpret every second ResNet block on level $l$ as a C-node. Consequently, when restricting the weights or gradients from level $l$ to $l-1$  we copy the corresponding parameters of each block.
When prolongating information upwards to a finer level, we copy the parameters $\theta_{i/2}^{l-1}$ for all even $i$'s on level $l$ and interpolate for all odd ones (Fig.~\ref{fig_transferops}).
Formally we write the restriction and prolongation as matrices $I_{l}^{l-1} \in \mathbb{R}^{K^{l-1} \times K^{l}}$, and $I_l^{l+1}  \in \mathbb{R}^{K^{l+1} \times K^{l}}$ respectively. $K^l$ and $K^{l-1}$ denote the number of parameters per level.
As smoother we use GD, which is available in all popular ML frameworks.

With this we can formulate sMG/OPT (Alg.~\ref{alg_smgopt}). The sMG/OPT optimizer loops over the mini-batches $j$ and applies one MG/OPT V-cycle.
The iteration starts at the finest level $L$. First, a coupling term $v^l$ is computed (lines 1-4), which stores the difference between the gradient on level $l$ and the projected gradient from level  $l+1$. 
For the  finest  level, $v^L$  is set to zero.
Then, $\nu^l$ GD steps are carried out (line 6) and the weights are restricted to the next lower level $l-1$.
This is followed by a recursive call to sMG/OPT (line 8), after which the correction $c^l$ is computed by prolonging the change of the weights in the lower level (line 9).
We carry out a line search using $c^l$ as direction and add $\alpha^l \,c^l$ to the weights on level $l$.
The iteration ends after an additional smoothing step.
At level $l=0$ we end the recursion, only smoothing is performed.

Note that the coupling term  $v^l$ plays a crucial role in sMG/OPT as it represents the "fine-to-coarse" defect \cite{Bra77}. 
It is used to add the additional term $<v^l,\theta_l>$ to the loss function at level $l$, in order to ensure 1st order consistency.
Hence the additional $v^l$ parameter in $\textrm{GD}(\theta^{l,0};l,j,\nu^l,v^l)$ on lines $6$,$11$, and $14$.

The line search in line~10 acts both as acceleration and convergence  control. Without it, sMG/OPT is equivalent to FAS. A stochastic variant of FAS has been explored in \cite{KGK20}.

\begin{algorithm}
\caption{sMG/OPT($l$, \ j, $\theta^{l,0}$) for level $l$, mini-batch $j$}\label{alg_smgopt}
\begin{algorithmic}[1]
\STATE \textbf{IF} $l=L$ 
\STATE $\quad$ $v^l = 0$
\STATE \textbf{ELSE} 
\STATE $\quad $ $v^l  = \nabla \ell(\theta^{l,0};\bar{x}_j,\bar{c}_j) - I_{l+1}^l \nabla \ell(\theta^{l+1,\nu^{l+1} };\bar{x}_j,\bar{c}_j)$
\STATE \textbf{IF} $l>0$ 
\STATE $\quad$ $\theta^{l,\nu^l} = \textrm{GD}(\theta^{l,0};l,j,\nu^l,v^l)$
\STATE $\quad$ $\theta^{l-1,0} = I_{l}^{l-1} \theta^{l,\nu^{l}}$
\STATE $\quad$ $\theta^{l-1,*}$ =  sMG/OPT($l-1$, j, $\theta^{l-1,0}$)
\STATE $\quad$ $c^l = I_{l-1}^l(\theta^{l-1,*}-\theta^{l-1,0})$
\STATE $\quad$ $\theta^{l,\nu^{l}+1} = \theta^{l,1} + \alpha^l c^l$ \textrm{ (line search)}
\STATE $\quad$ $\theta^{l,*} = \textrm{GD}(\theta^{l,\nu^l+1};l,j,\mu^l,v^l)$
\STATE $\quad$ \textbf{return} $\theta^{l,*}$
\STATE \textbf{ELSE} $l=0$
\STATE $\quad$ $\theta^{0,*} = \textrm{GD}(\theta^{0,1};l,j,\nu^0,v^0)$
\STATE $\quad$ \textbf{return} $\theta^{0,*}$
\end{algorithmic}
\end{algorithm}
\section{Numerical experiments}
We implemented sMG/Opt with the backtracking line search algorithm \cite{Noc06}~(Alg.~3.1) in the PyTorch framework \cite{PyTorch19}.
The code used for the experiments is publicly available, see~\cite{Pla+21}.

The efficiency of sMG/OPT was tested on the MNIST~\cite{LCB10} data set, the supplementary  material also contains tests on the MNIST1d~\cite{Gre20} data set.

The ResNets are set up  as described in Sec.~\ref{sec_formulation}, with the width $w$ fixed to $10$ for all layers ${n=1,\ldots,N}$.
At the beginning and the end, we have dense layers to map the input to the network width and the network width to the corresponding number of classes. 
The network depths ranged from $256$ to $2\,048$ residual blocks.
These networks do not reach SOA accuracy on MNIST ($92\%$ instead of the feasible $99+\%$). 
The aim was to have network architectures which have simple building blocks that are easy to combine to form large deep  networks and study their training. 

The learning rate was set to $0.1$, batch-size to $1,000$, and $\beta$ to $0$.
These values are close to the optimal SGD setup in order to compare the performance of SGD and sMG/OPT.
The number of pre- and post-smoothing steps $\nu^l$ $\mu^l$ is described in the supplementary material.
All configurations of SGD or sMG/OPT were run 5 times for a fixed number of epochs, with different but fixed random seeds.
The networks were initialized using Xavier initialization.

For the backtracking line search we used, if not mentioned otherwise, a starting value of $\bar{\alpha} =1$.  
This value was recursively shrunk by the factor $\rho =0.5$, until the Wolfe conditions were satisfied. 
We aborted the line search after $10$ steps if no appropriate value for $\alpha$ was found, and set $\alpha$ to $0$.
We note that adding line search increases the number of loss evaluations required by our method. 
However, as the loss is evaluated only on a mini-batch, this does not cause a rapid increase in the overall computational cost.

\subsection{Efficiency}
We trained for 5 epochs or 300 cycles (corresponding to the total number of mini-batches generated during 5 epochs). 
This was sufficient to reach top accuracy with the sMG/OPT configurations. 
Training to top accuracy with SGD would take longer, but  for clarity of presentation, we also stopped the training after 300 cycles.

\begin{figure}
    \centering
    \includegraphics[width=0.96\columnwidth]{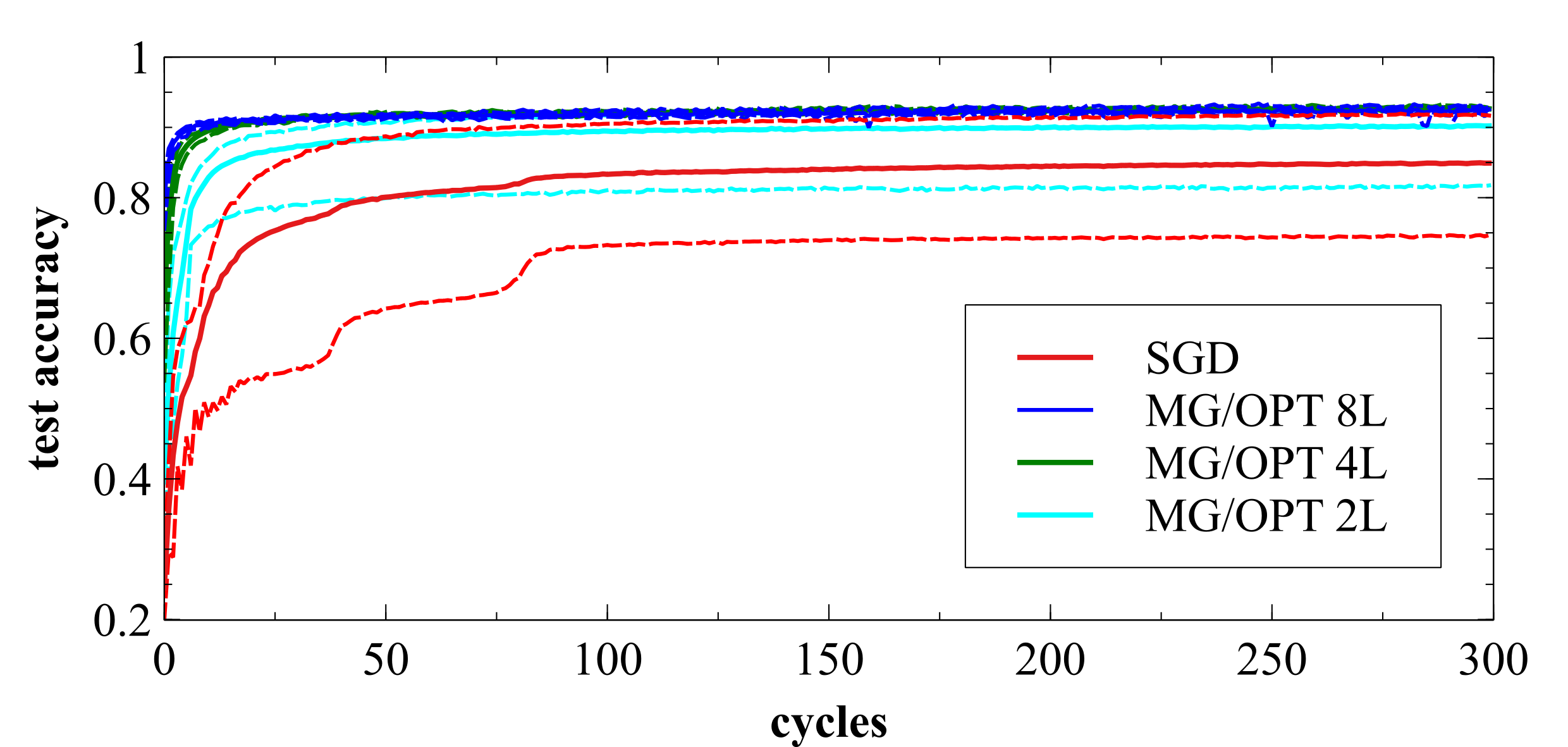}
    \caption{Convergence of different MG/Opt setups for MNIST from 1 (SGD) to 8 level methods. The ResNet contains $2\,048$ blocks. The dotted lines indicate +/- one standard deviation.}
    \label{fig_mnist_conv_levels}
\end{figure}

Fig.~\ref{fig_mnist_conv_levels} shows the  training of a ResNet with $2,048$ layers with SGD and sMG/OPT with $2$ to $8$ levels. 
We see that with sMG/OPT we can reach higher accuracies after fewer cycles, i.e. processing less data.
For example, after 50 cycles sMG/OPT with 8 levels has reached an accuracy of 91.7\% percent, whereas SGD has only reached 80\%.
Also, convergence of sMG/OPT is faster when using more levels with an attenuating effect when going from $4$ to $8$ levels.  
Interestingly, with respect to the different random seeds, training with sMG/OPT also leads to less variance. 
This suggests that the upward phase of sMG/OPT retains the variation reduction observed in the multilevel initialization schemes in \cite{CGS19, CMH+17}.

\begin{figure}
    \centering
    \includegraphics[width=0.96\columnwidth]{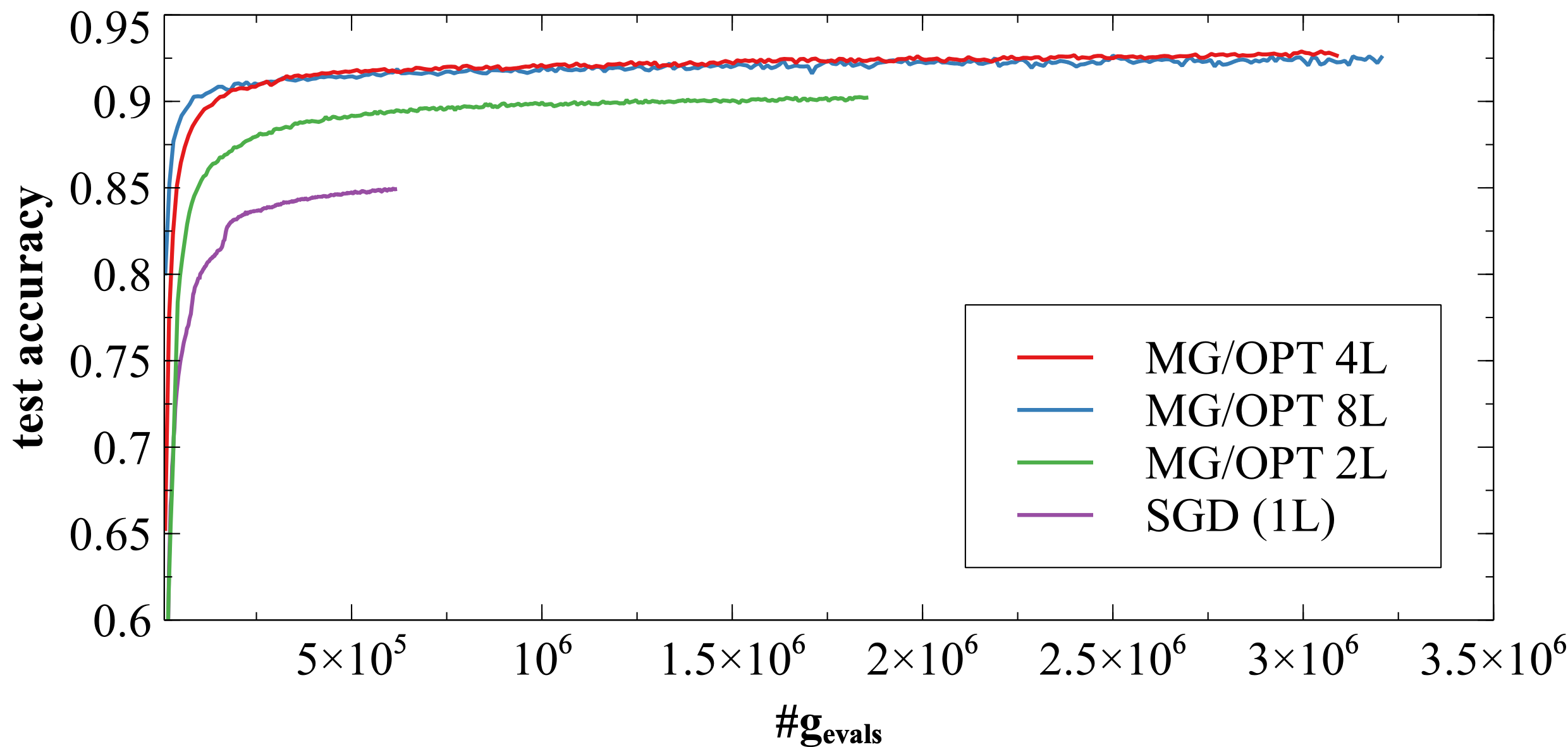}
    \caption{Convergence of MG/Opt with $2$,$4$,$8$ levels for MNIST with respect to the computational complexity.}
    \label{fig_mnist_work}
\end{figure}

However, the computational complexity of one sMG/OPT cycle is significantly higher than that of SGD.
To this purpose, we measure the computational cost by means of the number of required gradient evaluations, scaled by the number of layers of the network according to:
\begin{equation}
\label{eq_scaled_grad_evals}
    \# \textrm{g}_{\textrm{evals}} = \sum_{l=0}^L N^l Q^l,
\end{equation}
where $N^l$ and $Q^l$ stand for the number of layers and the number of gradient evaluations invoked on level $l$, respectively. The memory consumption of sMG/OPT is analogous.
For SGD we set $L$ to $0$.
Fig.~\ref{fig_mnist_work} shows the performance of the method  in terms of the number of required gradient evaluations $\#\textrm{g}_{\textrm{evals}}$.
Also from this view sMG/OPT reaches higher levels of accuracy with less overall work than SGD. 

\subsection{Evaluation of line search}
\begin{figure}
    \centering
    \includegraphics[width=0.96\columnwidth]{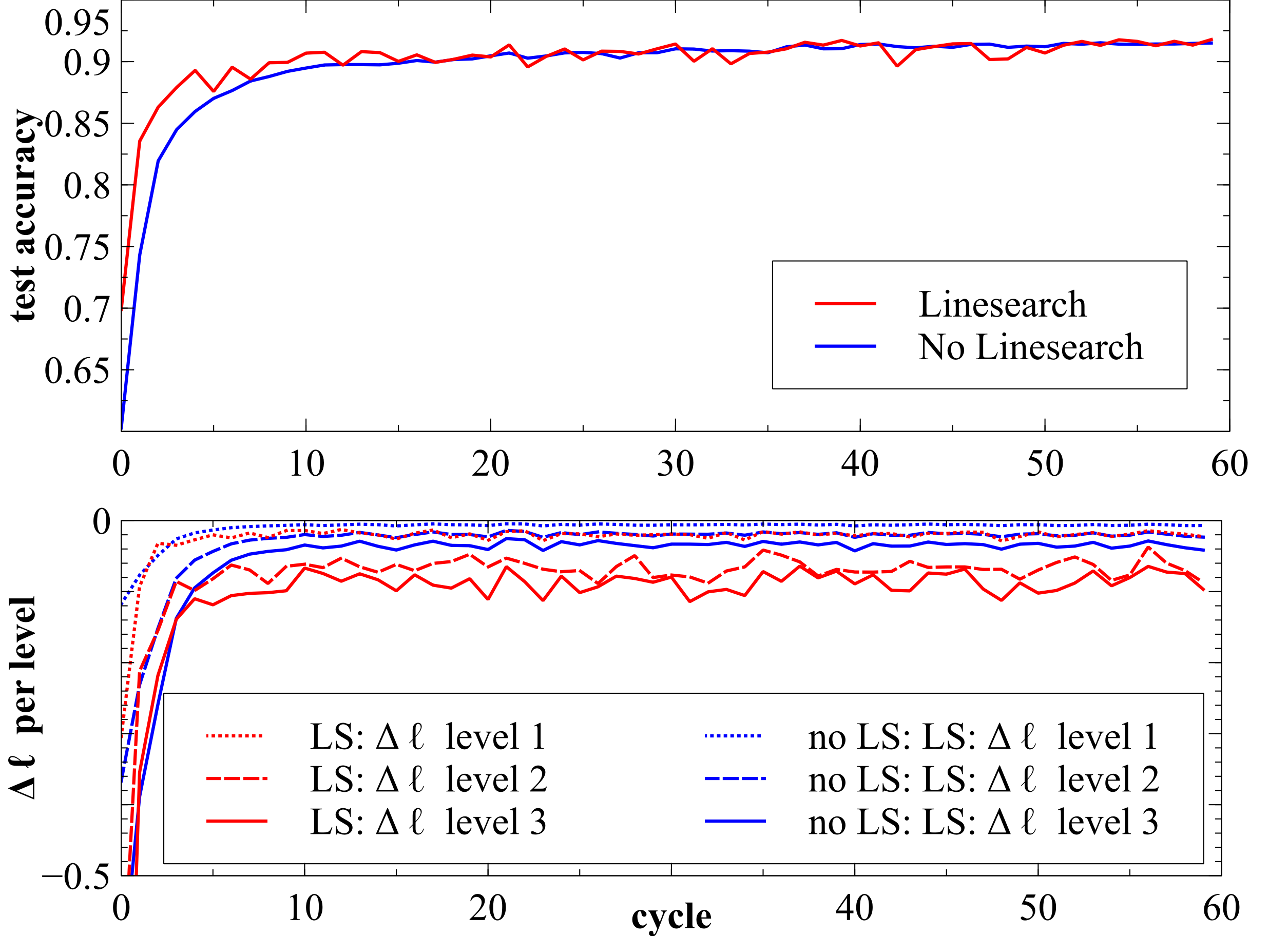}
    \caption{Top: Test accuracy of sMG/Opt with and without line search. Bottom: $\Delta \ell$ on each level for two runs with and without line search. Using an initial step-length of $3$ accelerates convergence in early stages.
    }
    \label{fig_linesearch}
\end{figure}

To investigate the effect of the line search, we trained a ResNet with $256$ blocks with a 4 level sMG/OPT method. Once with and once without line search, and set the initial step length $\bar{\alpha}$ to $3$ in order  to highlight the effect.
In Fig.~\ref{fig_linesearch}~(Top) we see that including a line search results in a speed-up in the early stages.
This speed-up can also be observed by looking at the changes of the loss functions at each level defined by 
$\Delta \ell^l = \ell^l(\theta^l;x,c) - \ell^l(\theta^{l} + \alpha^l c^l;x,c)$.

These differences are depicted in Fig.~\ref{fig_linesearch}~(bottom) and show that the line search gives an extra bit of efficiency at every cycle with consistently lower values of $\Delta \ell^l$. 
These bigger reductions in loss at all levels of the sMG/OPT cycle then yield faster convergence.
Another reason for employing the line search algorithm is to nullify spurious corrections from the lower levels, that is, to act as convergence control. To this we refer to  the supplementary material.

\subsection{Accuracy of the auxiliary networks}
In an sMG/OPT training scheme with $L+1$ levels, the auxiliary networks, associated with levels $0$ to $L-1$ are trained to similar accuracies and can be used for inference.
Table~\ref{tab_accs_pruning} depicts the test accuracies of ResNets associated with every second level of the multilevel hierarchy after $300$ cycles with 8-level sMG/OPT. 
The reported results are not necessarily the top accuracies, since we have not used early stopping.  
We observe that for $2\,048$ blocks the test accuracies are approximately the same, down to the level 3. 
The same applies to the ResNets with $1\,024$ blocks, whereas for depths of $256$ and $512$, the test accuracies deteriorate earlier. 
It should be noted that for a ResNet of depth $2\,048$, the auxiliary ResNet at level 3 has a depth of $128$ blocks and this auxiliary network has the same or even higher accuracy than the networks with $256$ and $512$ blocks at their respective top-level. 
This indicates that sMG/Opt can also be used as a pruning technique to yield smaller, but more accurate networks.

\begin{table}
\centering
\begin{tabular}{|c|c|c|c|c| }
\hline
\multirow{2}{*}{Level} & \multicolumn{4}{|c |}{ $\#$ ResNet blocks} \\
\cline{2-5}
 & $256$ & $512$ & $1\,024$  & $2\,048$  \\
\hline
7 & 91.8 & 90.3 & 92.4 & 92.5 \\
5 & 92.0 & 90.2 & 92.5 & 92.5 \\
3 & 86.9 & 88.0 & 91.2 & 92.3 \\
1 & 58.4 & 63.0 & 82.0 & 84.3 \\
0 & 72.7 & 76.2 & 75.6 & 70.1  \\
\hline
\end{tabular}
\caption{Test accuracies of the ResNets associated with different levels of sMG/Opt at the end of $300$ cycles.}
\label{tab_accs_pruning}
\end{table}

\section{Discussion}
We extended MG/OPT to the stochastic setting as multigrid methods are attractive candidates to train deep neural networks.
The experiments on ResNets with up to $2,048$ layers showed that training with sMG/OPT can be more efficient than with the corresponding SGD variants.
Furthermore, the performance of sMG/OPT method is less susceptible to the choice of different random seeds, as shown by the lower standard deviations of the convergence plots.
We also investigated how the line search accelerates the training, by allowing bigger step-lengths for the coarse grid corrections. 
Lastly, training with sMG/OPT has potential as a pruning technique.
The top $4$ levels in sMG/OPT training, configured with $8$ levels, yield about the same accuracy as the deepest network, which indicates that these networks can be used for computationally cheaper inference. 

Naturally, the method now needs to be extended to train more common data sets, e.g.,~CIFAR-10, CIFAR-100, or ImageNet, and with more common architectures such as CNNs. This is the focus of our ongoing and future work.

\section*{Appendix}
\appendix
\label{sec_appendix}
In the appendix we provide the information referred to as supplementary material, but also selected additional material. In particular, we show the speed-up with respect to wall time, results from alternative MG/OPT configurations with different number of pre- and postsmoothing steps, results from training with the MNIST1d data set, and give a deeper look into the line search step of sMG/OPT.

\section{sMG/OPT configurations}
Table~\ref{tab_cfg_mnist_1} shows the sMG/OPT configurations in the paper, while Table~\ref{tab_cfg_mnist_2} shows alternative sMG/OPT configurations used to train MNIST and MNIST1d.

The numbers in the second row denote the number of levels of a given sMG/OPT configuration, whereas the values $(\nu,\mu)$ in each row denote the number of pre- and post-smoothing steps on the level denoted in the left column. At the finest and at the coarsest level, we used no post-smoothing steps.

\begin{table}[h]
    \centering
    \begin{tabular}{|c|c|c|c|}
    \hline
    \multirow{2}{*}{Level} & \multicolumn{3}{c|}{ \#Levels in sMG/Opt} \\
    \cline{2-4}
    & 2 & 4 & 8 \\
    \hline
    0 & (2,0) & (2,0) & (2,0) \\
    1 & (1,0) & (2,2) & (2,2) \\
    2 &     &  (1,1) & (2,2) \\
    3 &     & (1,0) & (2,2) \\
    4 &     &  & (1,1) \\
    5 &     &  & (1,1) \\
    6 &     &  & (1,1) \\
    7 &     &  & (1,0) \\
    \hline
    \end{tabular}
    \caption{sMG/OPT pre- and postsmoothing steps configurations in article.}
    \label{tab_cfg_mnist_1}
\end{table}

\begin{table}[h]
    \centering
    \begin{tabular}{|c|c|c|c|}
    \hline
    \multirow{2}{*}{Level} & \multicolumn{3}{c|}{ \#Levels in sMG/Opt} \\
    \cline{2-4}
    & 2 & 4 & 8 \\
    \hline
    0 & (1,0) & (1,0) & (1,0) \\
    1 & (1,0) & (1,1) & (1,1) \\
    2 &     &  (1,1) & (1,1) \\
    3 &     & (1,0) & (1,1) \\
    4 &     &  & (1,1) \\
    5 &     &  & (1,1) \\
    6 &     &  & (1,1) \\
    7 &     &  & (1,0) \\
    \hline
    \end{tabular}
    \caption{Alternative sMG/OPT pre- and postsmoothing steps configurations for MNIST.}
    \label{tab_cfg_mnist_2}
\end{table}

\section{MNIST}
\subsection{Speed-up with respect to wall time}
\begin{figure}[h]
    \centering
    \includegraphics[width=0.96\columnwidth]{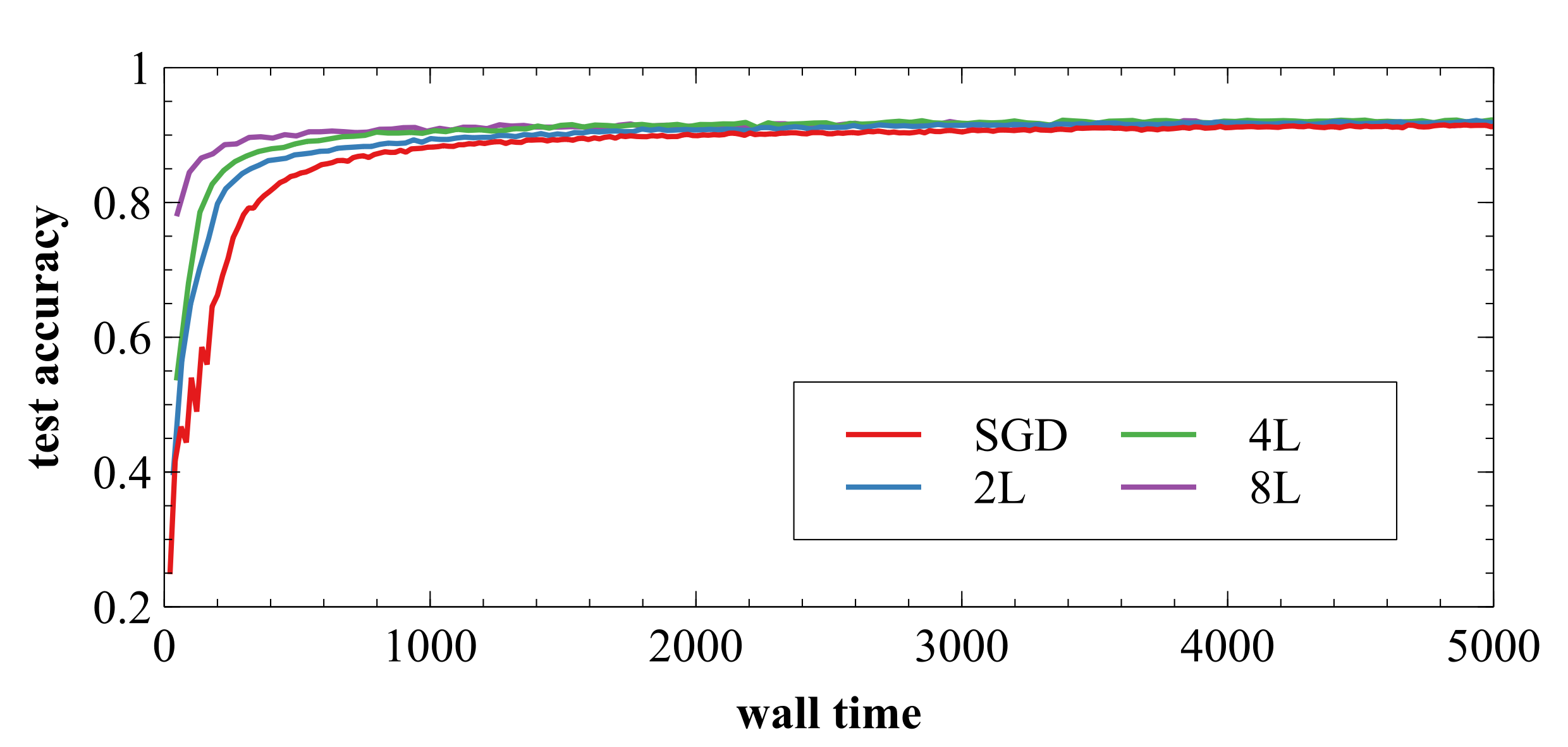}
    \caption{Speed-up of sMG/OPT with respect to wall time.}
    \label{fig_mnist_time}
\end{figure}

Despite the high-level implementation of sMG/OPT in PyTorch, we were able to achieve a speed-up with respect to wall-time, albeit less pronounced, which is shown in Fig.~\ref{fig_mnist_time}. 
We believe that with a low-level implementation of the transfer operators, the speed-up could be further improved.

\subsection{Convergence behavior for fine level with $1,024$ layers}
\begin{figure}[h]
    \centering
    \includegraphics[width=0.96\columnwidth]{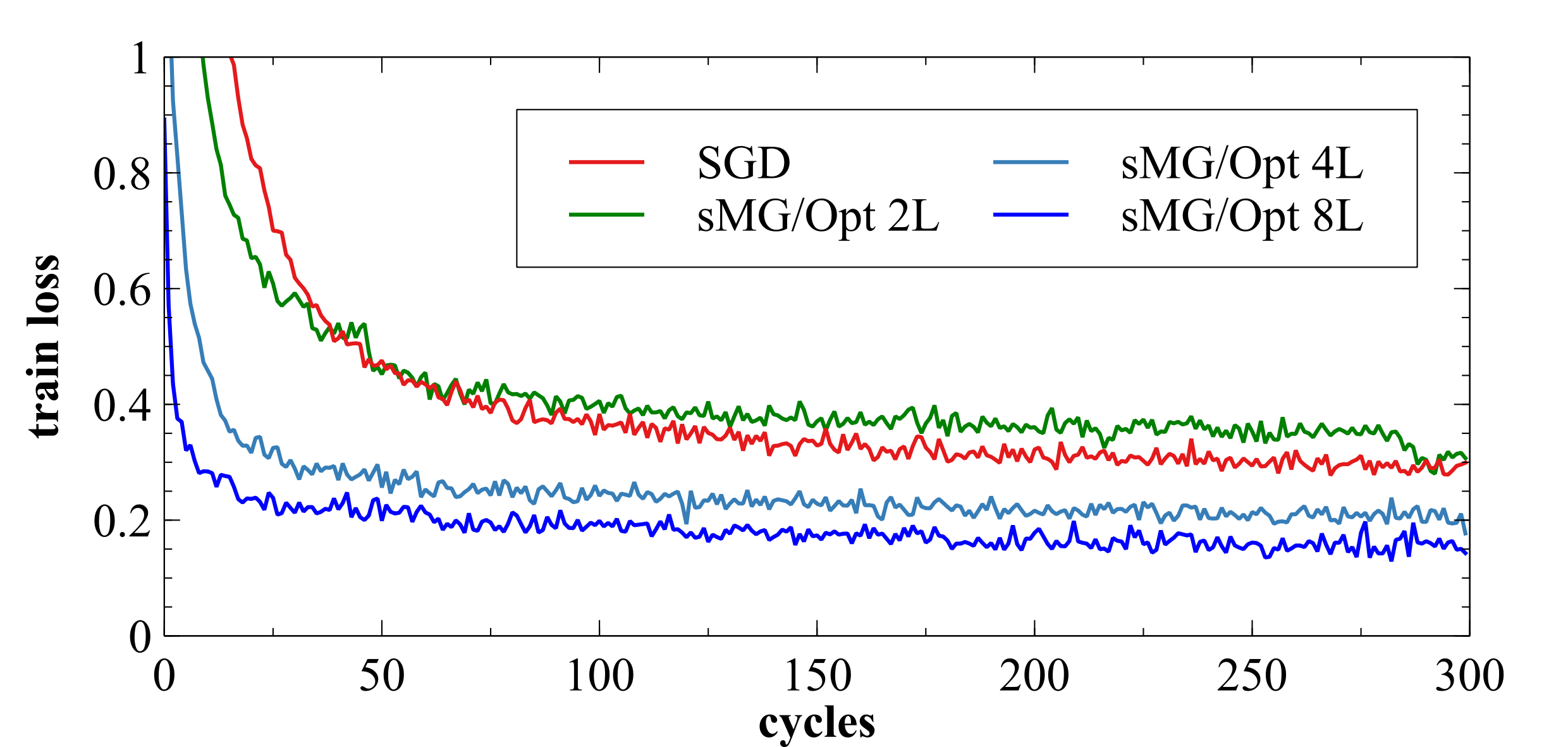}
    \includegraphics[width=0.96\columnwidth]{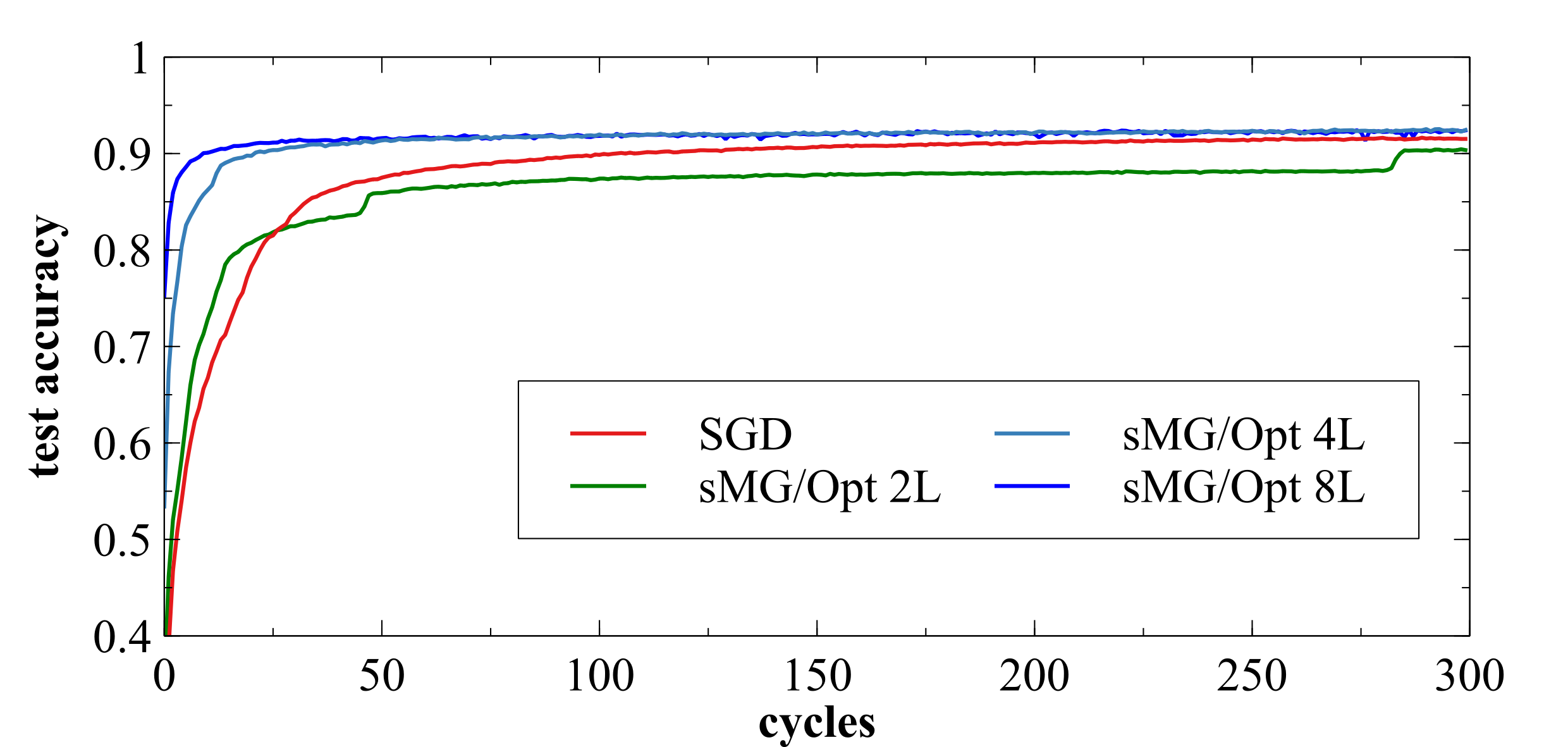}
    \caption{Train loss (top) and test accuracy per sMG/Opt configuration from Table~\ref{tab_cfg_mnist_1} for a ResNet with $1024$ blocks.}
    \label{fig_mnist_1024}
\end{figure}

To underline that multilevel speed-up behaves similarly for different fine level resolutions, we show in Fig.~\ref{fig_mnist_1024} simulations with $1\,024$  block ResNets with the configuration of Table~\ref{tab_cfg_mnist_1}. 
We plot the training loss (top) and the test accuracy (bottom) with respect to the number of cycles. 
The results are averaged over $5$ runs. 
Like the $2\,048$ block case, convergence speed correlates with the number of levels in the first cycles. 
However, in the case of the $2$-level setup, we observe poorer convergence and top accuracy, which is due to one outlier in the $5$ simulations. 
This again suggests that the multilevel training is more robust for deeper networks and also that the overall training scheme is more robust for $4$ and more levels.

\subsection{Pre- and postsmoothing steps}
\begin{figure}[h]
    \centering
    \includegraphics[width=0.96\columnwidth]{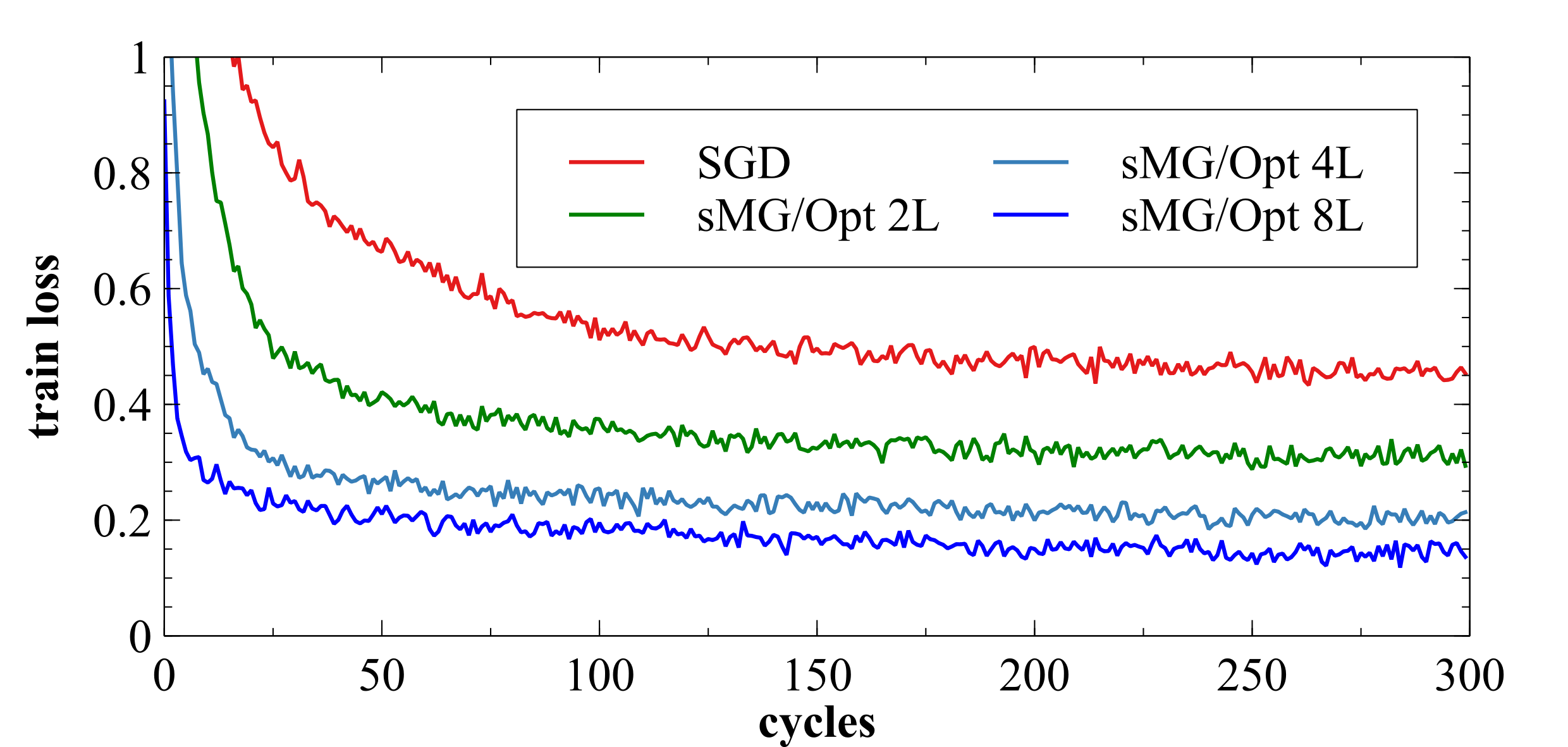}
    \includegraphics[width=0.96\columnwidth]{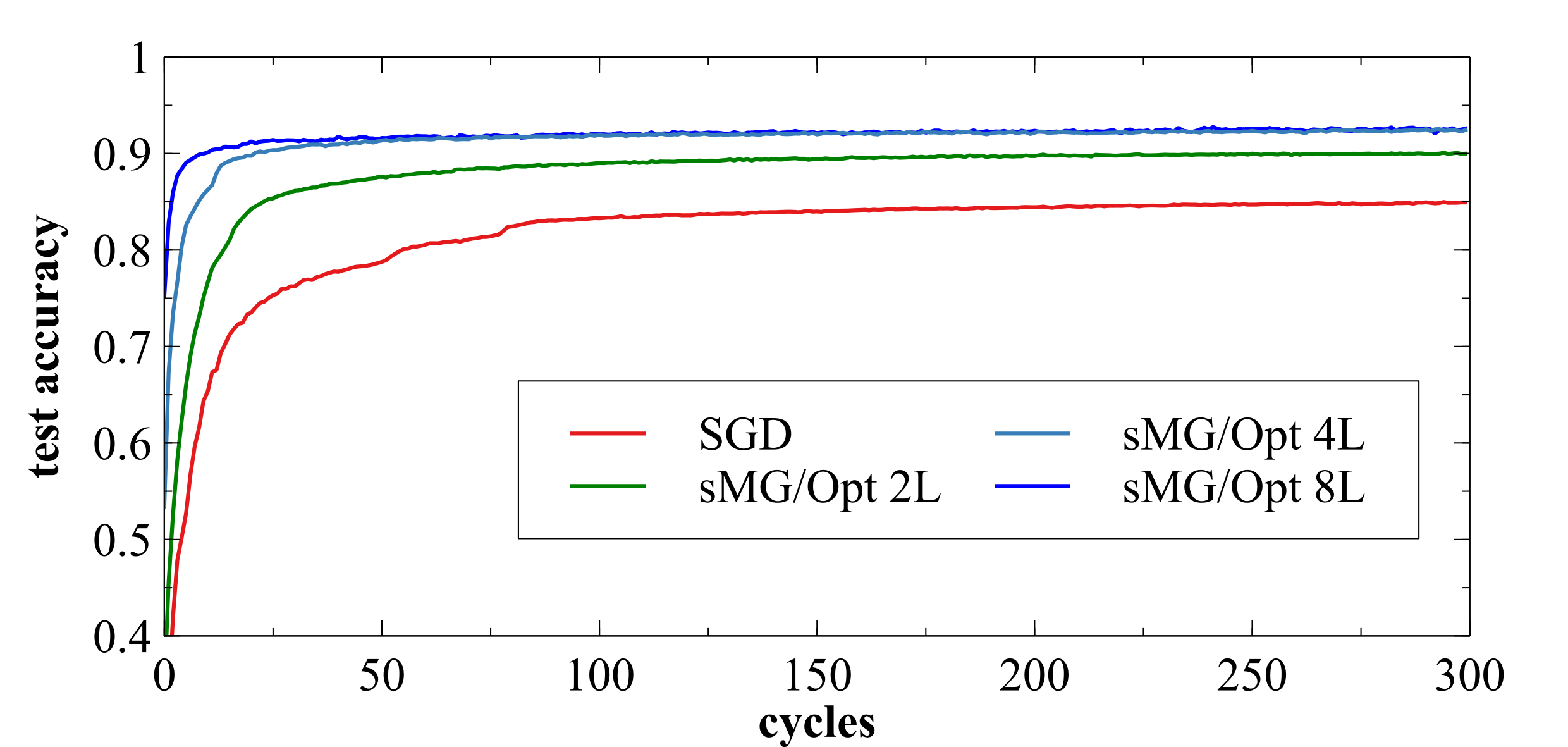}
    \caption{Train loss (top) and test accuracy per sMG/OPT configuration from Table~\ref{tab_cfg_mnist_2} for a ResNet with $2048$ blocks.}
    \label{fig_mnist_2048}
\end{figure}

Not shown in the article are all the setups of sMG/OPT for MNIST with the configurations in Table~\ref{tab_cfg_mnist_2}, where we ran the simulations with just one pre- and postsmoothing step.
As a representative for all these simulations, we show  in Fig.~\ref{fig_mnist_2048} the convergence for a ResNet with $2\,048$ residual blocks. 
Again, we show the training loss (top) together with the test accuracy (bottom). 
We observe that the speed-ups are qualitatively the same, the only difference being a more pronounced differentiation between the $4$ and $8$ level sMG/OPT setup.
Last but not least, we like to mention that these speed-ups were also observed for much shallower networks with $8$ or $16$ ResNet blocks. 
These are not shown here, as our focus was to study the multilevel training of (pathologically) deep ResNets.

\begin{figure}[h]
    \centering
    \includegraphics[width=0.96\columnwidth]{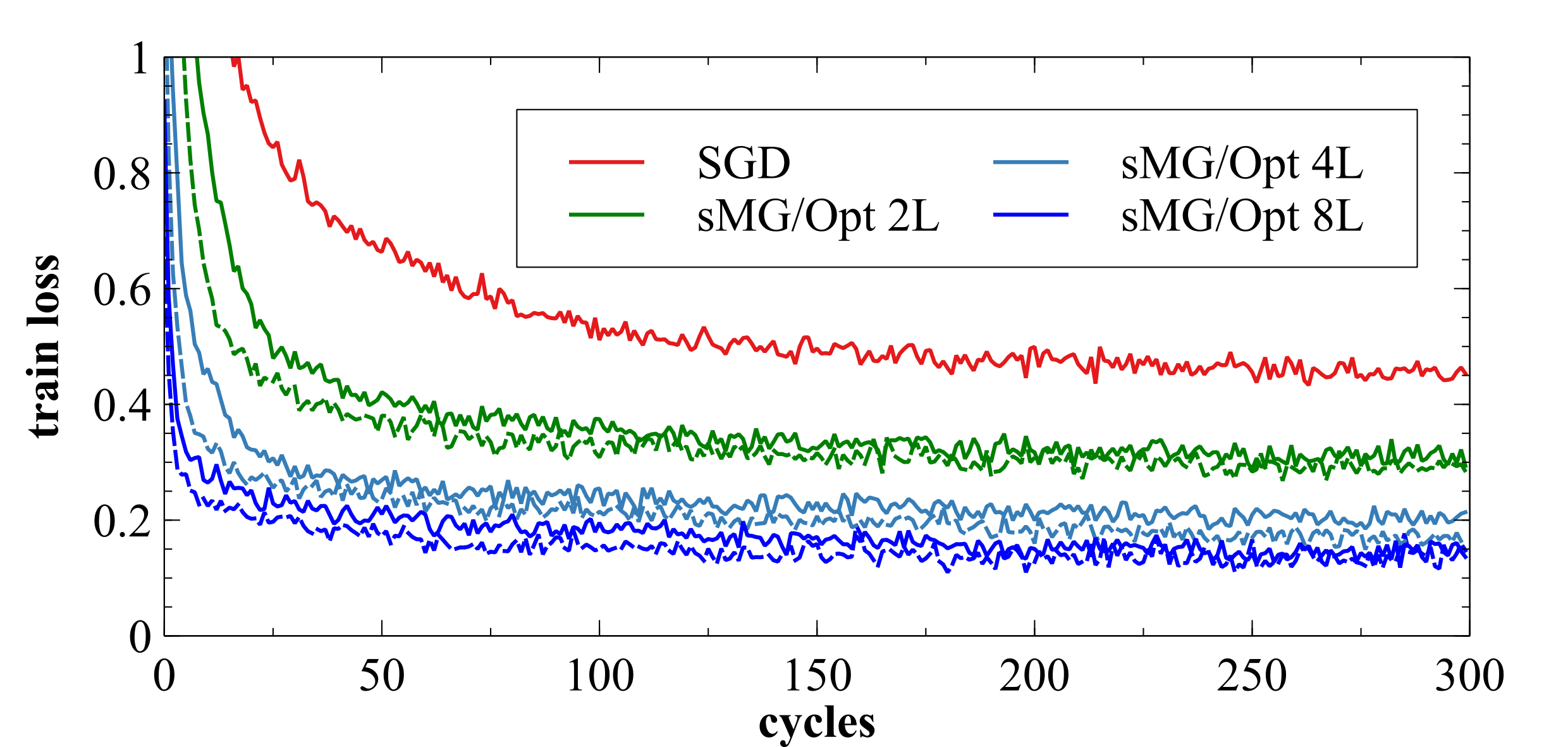}
    \includegraphics[width=0.96\columnwidth]{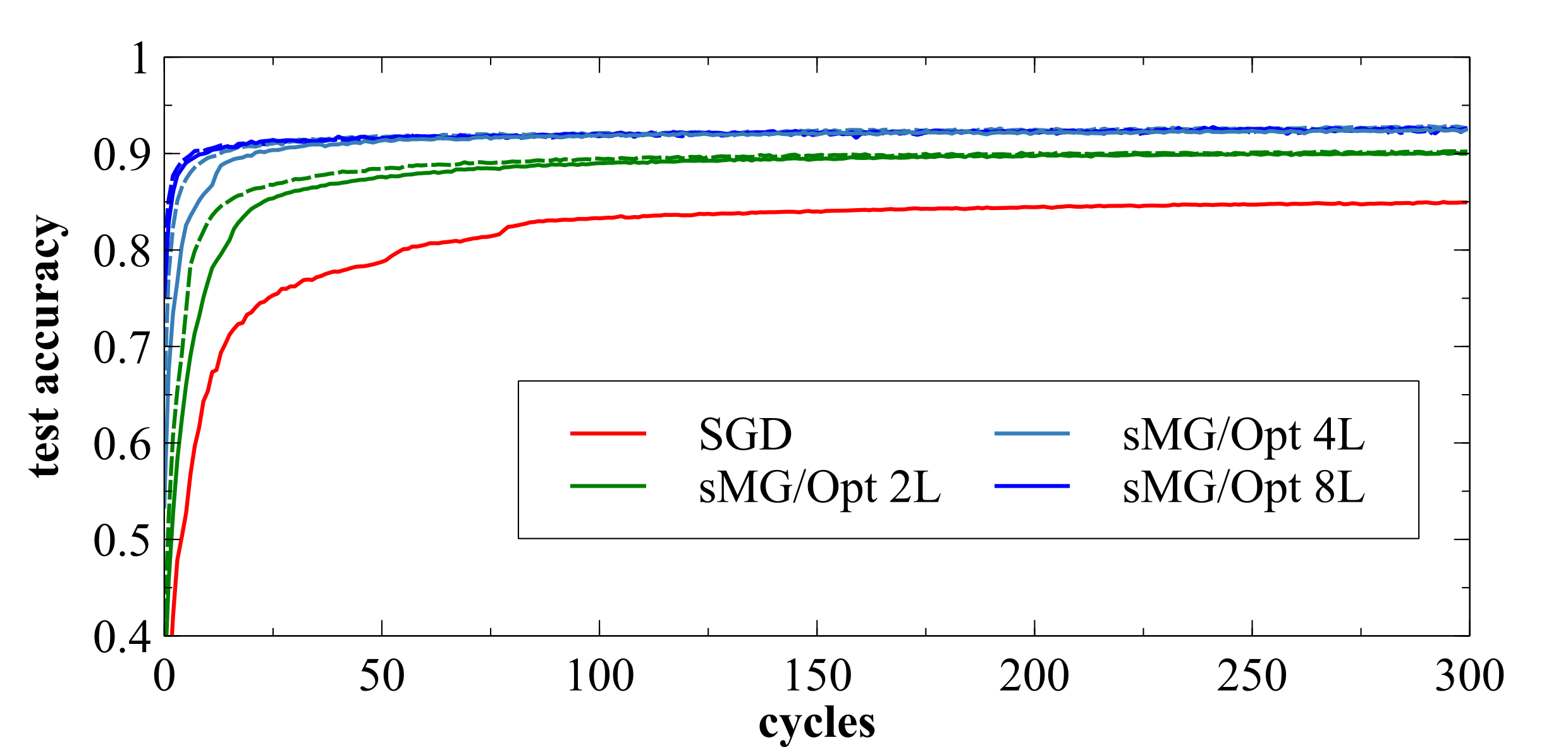}
    \caption{The influence of the number of pre- and post-smoothing steps.  sMG/OPT performance with the setup of Tab~\ref{tab_cfg_mnist_1} (dotted lines) versus the setup of Tab~\ref{tab_cfg_mnist_2} (solid lines).}
    
    \label{fig_mnist_comparison}
\end{figure}

We were also interested in the relative performance of the sMG/OPT setups from Table~\ref{tab_cfg_mnist_1} versus the ones in Table~\ref{tab_cfg_mnist_2}.
This is depicted in Fig.~\ref{fig_mnist_comparison}, where we plotted the training loss and test accuracy of both setups together. 
The simulations with the setups used in the article are plotted with dotted lines and those with the setup from Tab.~\ref{tab_cfg_mnist_2} with solid lines. 
We see that the additional post- and pre-smoothing steps do further increase the convergence speed, however, further hyper-parameter searches would have to be conducted to derive the optimal setup.

Lastly, to make the improvements in convergence speed more quantifiable, we summarize the test accuracies of the simulations shown in Fig.~\ref{fig_mnist_2048} in Table~\ref{tab_test_accs_cycles} after $5$, $10$, $50$, $100$, and $300$ cycles.
From the table it becomes evident that the biggest speed-ups occur within the first 50 cycles. We like to note here, that with the  $8$-level setup we reach a test accuracy of $90$ percent after $10$ cycles, while having only processed $10\,000$ of the $60\,000$ samples. It remains to be seen whether this is an oddity related to the MNIST data set, or whether the early "predictive powers" are an inherent part of sMG/OPT training (the former seems to be the case as we do not observe this phenomenon with MNIST1d).

\begin{table}
\centering
\begin{tabular}{|c|c|c|c|c|c| }
\hline
\multirow{2}{*}{sMG/OPT levels} & \multicolumn{5}{|c |}{ $\#$ Cycles} \\
\cline{2-6}
 &$5$& $10$ & $50$  & $100$ & $300$  \\
\hline
1 & 52.7 & 65.4 & 78.9 & 83.3 & 85.0\\
2 & 66.0 & 76.7 & 87.6 & 89.0 & 90.0 \\
4 & 83.5 & 86.8 & 91.6 & 91.9 & 92.6 \\
8 & 89.1 & 90.1 & 91.6 & 92.0 & 92.4 \\
\hline
\end{tabular}
\caption{Test accuracies of the different sMG/OPT setups after $5$,$10$, $50$, $100$,and $300$ cycles for training MNIST with a ResNet of $2048$ blocks.}\label{tab_test_accs_cycles}
\end{table}
\section{MNIST1d}
\begin{figure}
    \centering
    \includegraphics[width=0.96\columnwidth]{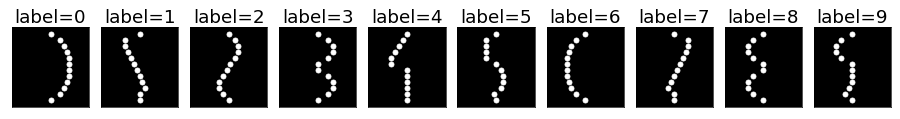}
    \includegraphics[width=0.96\columnwidth]{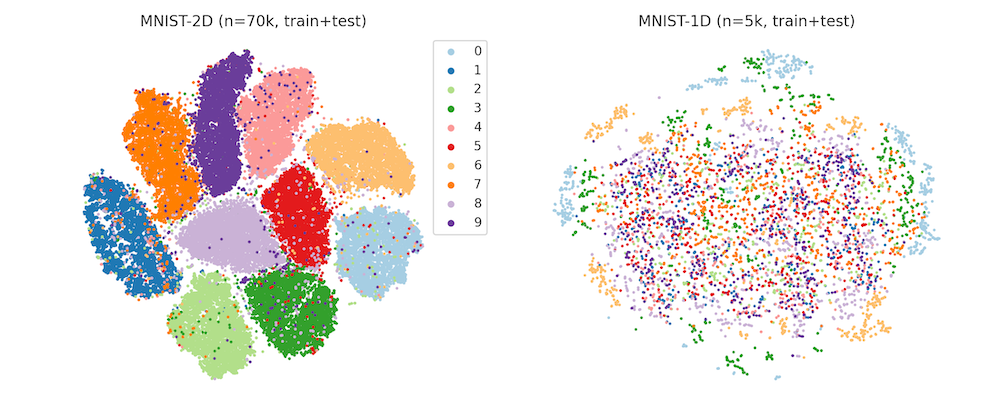}
    \caption{MNIST1d. Top: Prototypes of the $10$ classes, out of which the train and test samples are generated. 
    Bottom: MNIST and MNIST-1D data sets plotted with tSNE. 
    The clusters in MNIST suggest that MNIST is separable via a kNN classifier in pixel space. 
    Source of the picture~\cite{Gre20}.}
    \label{fig_mnist1d}
\end{figure}

\begin{figure}[h]
    \centering
    \includegraphics[width=0.96\columnwidth]{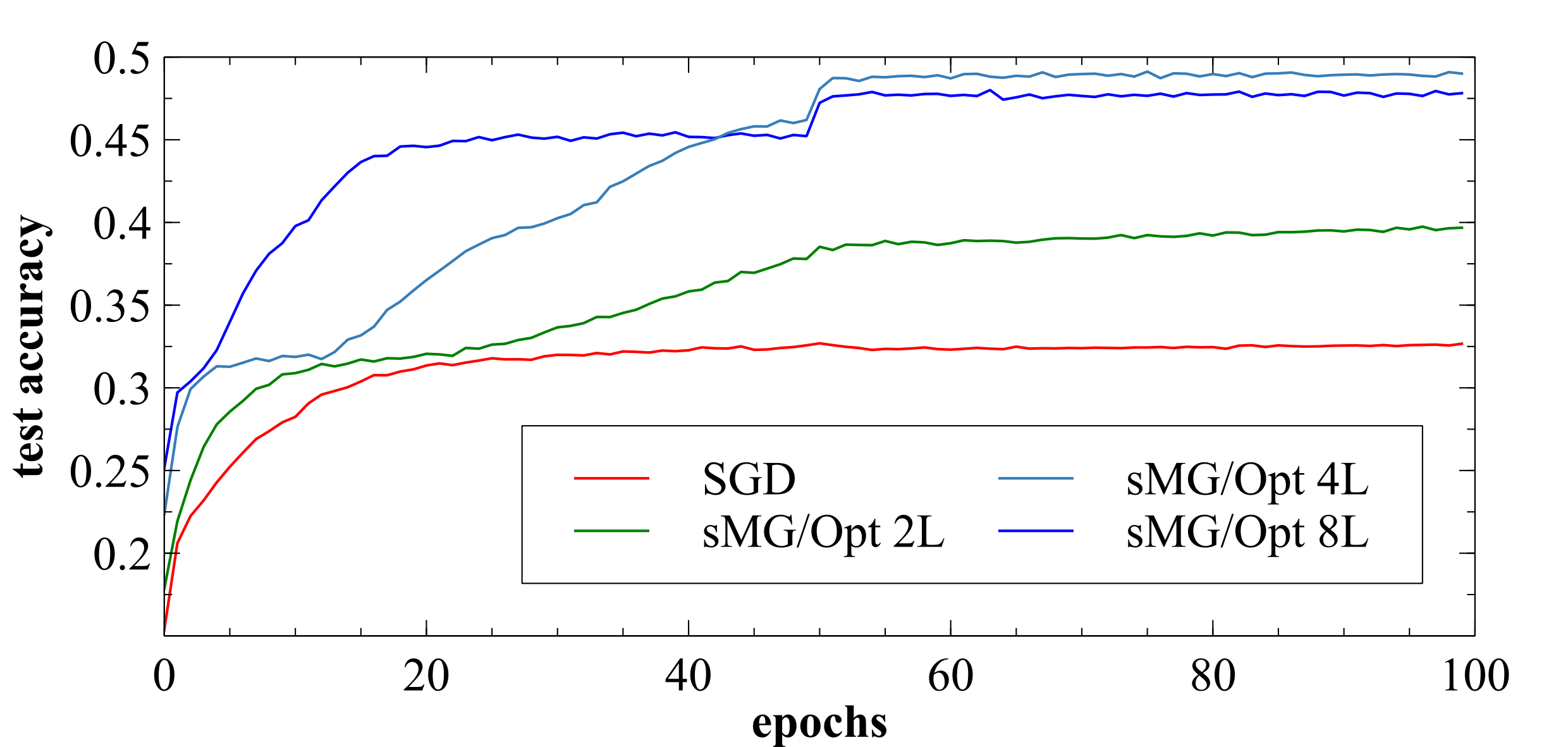}
    \caption{Training MNIST1d with a residual network of $512$  blocks and sMG/Opt configured for $2$,$4$,and $8$ levels.}    
    \label{fig_mnist1d_512}
\end{figure}

MNIST1d \cite{Gre20} is a new data set, designed to be small, but difficult to train.
It is similar to MNIST as it contains symbols loosely representing numbers. 
The samples are artificially generated from 10 template symbols, having the  dimension $x^i \in \mathbb{R}^{40}$. 
They are designed to be less well linearly separable than MNIST (Fig.~\ref{fig_mnist1d}).
The training and test data sets contain $4\,000$ and $1\,000$ samples each. 
Apart from being small, MNIST1d was also designed to differentiate more between different network architectures than MNIST. 
For example, training MNIST1d with a multilayer perceptron yields a test accuracy of $68\%$, while training it with a convolutional network yields $94\%$ (we can obtain $99\%$ in both cases for MNIST). 
\begin{figure}
    \centering
    \includegraphics[width=0.96\columnwidth]{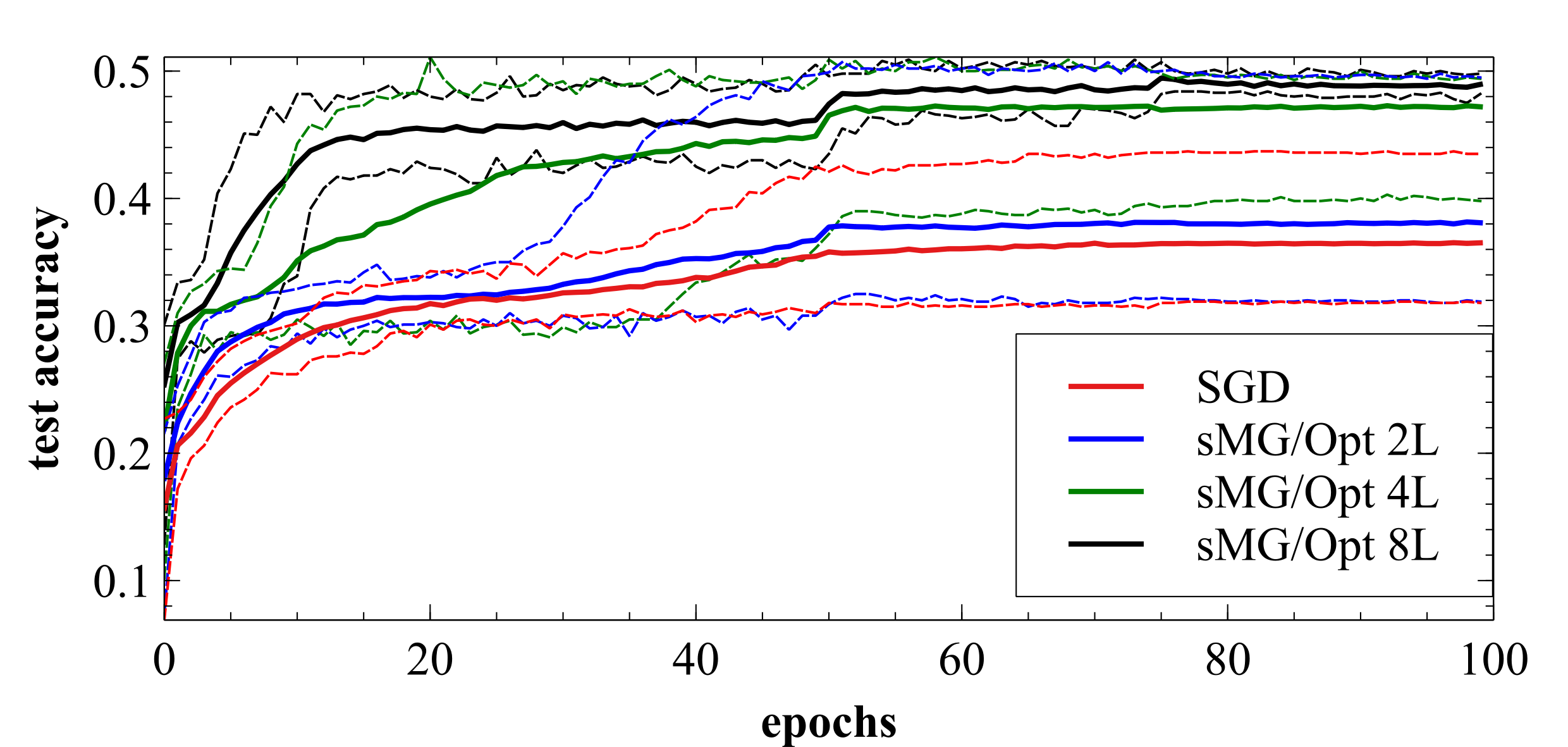}
    \caption{Training MNIST1d with a residual network of $256$ blocks and SMG/OPT configured for 2,4,and  8 levels. The dotted lines indicate one standard deviation w.r.t to the test accuracy.}
    \label{fig_mnists1d_epochs}
\end{figure}
In Figure~\ref{fig_mnist1d_512} we show the convergence of training MNIST1d sMG/OPT on a ResNet with $512$  blocks.
The results show the averages of $5$ runs, however this time we omit plotting the standard deviations.
We see that the convergence for this ResNet geometry follows qualitatively the same pattern, with the exception that the $4$-level setup eventually shows a higher top accuracy than the $8$-level setup.
This is possibly a consequence of the nature of the difficult to train MNIST1d data set, i.e. that significant randomness persists in the results even after averaging over $5$ runs.
While we do not plot all the individual runs here, we can summarize that in these simulations sMG/OPT optimizers were always outperforming SGD, and that the performance of the $4$-level sMG/OPT setup seems to perform as good, and sometimes better, as the $8$-level setup.

\section{Line search}
As mentioned in the article, the line search also adds additional robustness to the training procedure.
To illustrate this, we set up a simulation on a ResNet with $256$ blocks, where we used learning rates of $0.2$ and $\bar{\alpha}=2$, while the number of pre- and post-smoothing steps was as shown in Tab.~\ref{tab_cfg_mnist_1}.
With such a configuration, the correction steps $c^l$  (see Alg.~1) from the lower levels in the multilevel hierarchy are more prone to become poorly aligned to the finer level and start to hamper the performance of the training procedure.

\begin{figure}[h]
    \centering
    \includegraphics[width=0.96\columnwidth]{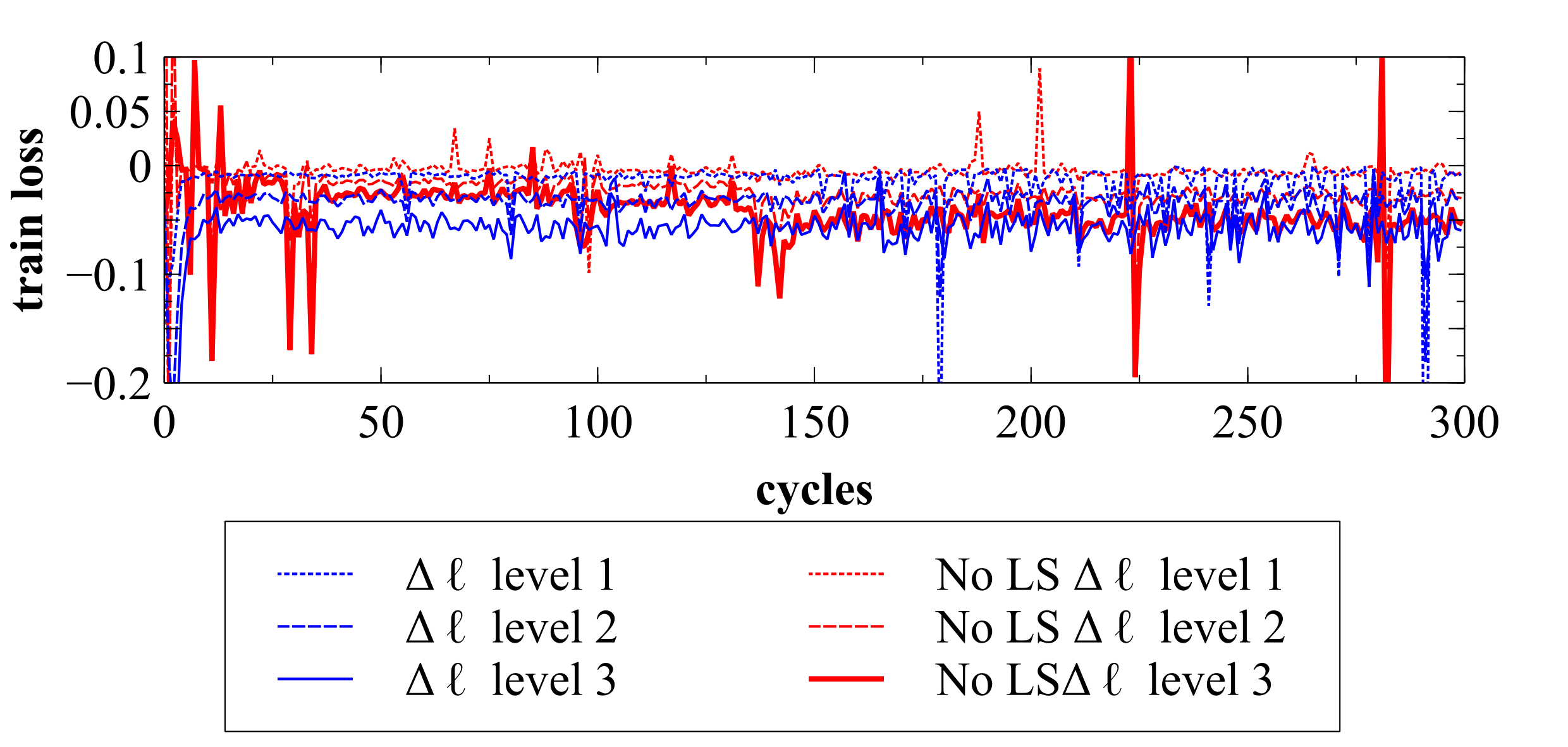}
    \includegraphics[width=0.96\columnwidth]{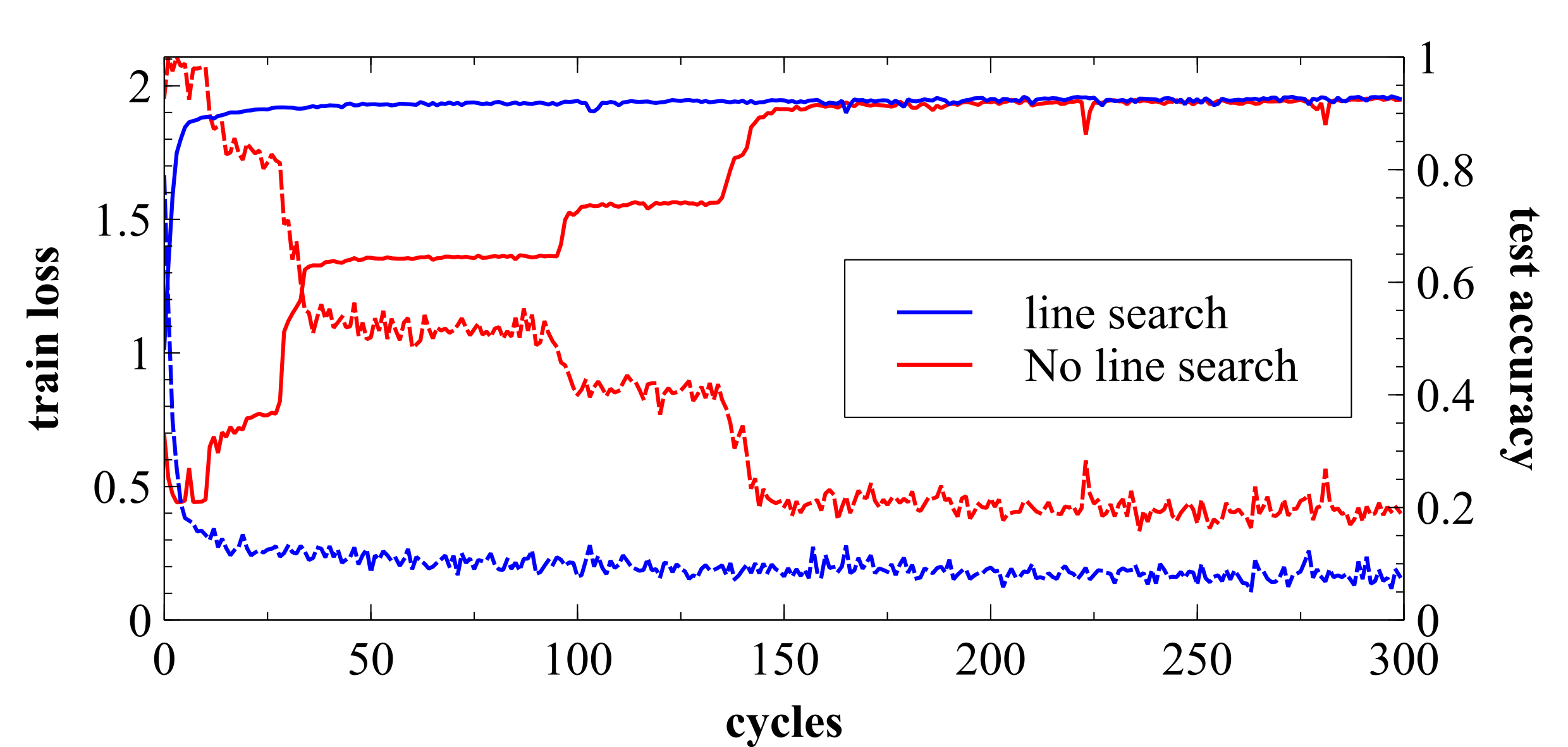}
    \caption{$4$-level sMG/OPT applied to a $256$ block ResNet. Blue: with line search, red: without line search.}   
    \label{fig_ls_appendix}
\end{figure}

We see this in Fig.~\ref{fig_ls_appendix}, where we show two $4$-level sMG/OPT simulations, once run with line search (blue lines), and once run without line search (red lines). 
The figure at the top  shows the $\Delta \ell$ for all the levels, and we can observe that the the non-line search variant produces spurious coarse grid corrections with $\Delta \ell>0$ throughout the $300$ cycles shown here.
The line search variant on the other hand, is able to control the updates, leading to $\Delta \ell \leq0$. 
This in turn leads to a faster and more steady convergence.
The test accuracy itself at the end of the 300 cycles does not seem to be affected by this, however, the continuing incorporation of "bad" corrections eventually shows up in the overall training loss, which remains higher than in the variant with line search.

\section{Hardware}
The experiments were run on the cluster of our institute (CPU: 2.30GHz Intel Xeon E5-2650 v3, GPU: NVIDIA GeForce GTX 1080).

\bibliography{central_biblio, biblio_minipaper}
\bibliographystyle{icml2021}

\end{document}